\def\year{2021}\relax
\documentclass[letterpaper]{article} 
\usepackage{aaai21}  
\usepackage{times}  
\usepackage{helvet} 
\usepackage{courier}  
\usepackage[hyphens]{url}  
\usepackage{graphicx} 
\usepackage{natbib}  
\usepackage{caption} 
\usepackage{amsmath}
\usepackage{bm}
\usepackage{amsfonts,amssymb}

\newcommand{\etal}{\textit{et al}.~}
\usepackage{array}
\usepackage{booktabs}
\usepackage{multirow}
\usepackage[ruled,vlined]{algorithm2e}
\usepackage[switch]{lineno}
\nocopyright
\title{Semi-Supervised Learning for In-Game Expert-Level Music-to-Dance Translation}

\author{
Yinglin Duan,\textsuperscript{\rm 1} \thanks{These authors contributed equally to this work}
Tianyang Shi,\textsuperscript{\rm 1} $^*$
Zhengxia Zou,\textsuperscript{\rm 2} 
Jia Qin,\textsuperscript{\rm 1, \rm 3}
Yifei Zhao,\textsuperscript{\rm 1}
Yi Yuan,\textsuperscript{\rm 1}\thanks{Corresponding author: yuanyi@corp.netease.com}\\
Jie Hou,\textsuperscript{\rm 1}
Xiang Wen,\textsuperscript{\rm 1, \rm 3}
Changjie Fan\textsuperscript{\rm 1}\\ 
\textsuperscript{\rm 1} NetEase Fuxi AI Lab ~
\textsuperscript{\rm 2} University of Michigan, Ann Arbor ~
\textsuperscript{\rm 3} Zhejiang University ~
}

\begin{document}

\newcolumntype{L}[1]{>{\raggedright\arraybackslash}p{#1}}
\newcolumntype{C}[1]{>{\centering\arraybackslash}p{#1}}
\newcolumntype{R}[1]{>{\raggedleft\arraybackslash}p{#1}}

\maketitle

\begin{abstract}
Music-to-dance translation is a brand-new and powerful feature in recent role-playing games. Players can now let their characters dance along with specified music clips and even generate fan-made dance videos. Previous works of this topic consider music-to-dance as a supervised motion generation problem based on time-series data. However, these methods suffer from limited training data pairs and the degradation of movements. This paper provides a new perspective for this task where we re-formulate the translation problem as a piece-wise dance phrase retrieval problem based on the choreography theory. With such a design, players are allowed to further edit the dance movements on top of our generation while other regression based methods ignore such user interactivity. Considering that the dance motion capture is an expensive and time-consuming procedure which requires the assistance of professional dancers, we train our method under a semi-supervised learning framework with a large unlabeled dataset (20x than labeled data) collected. A co-ascent mechanism is introduced to improve the robustness of our network. Using this unlabeled dataset, we also introduce self-supervised pre-training so that the translator can understand the melody, rhythm, and other components of music phrases. We show that the pre-training significantly improves the translation accuracy than that of training from scratch. Experimental results suggest that our method not only generalizes well over various styles of music but also succeeds in expert-level choreography for game players.
\end{abstract}

\section{Introduction}

\begin{figure}
  \includegraphics[width=0.9\linewidth]{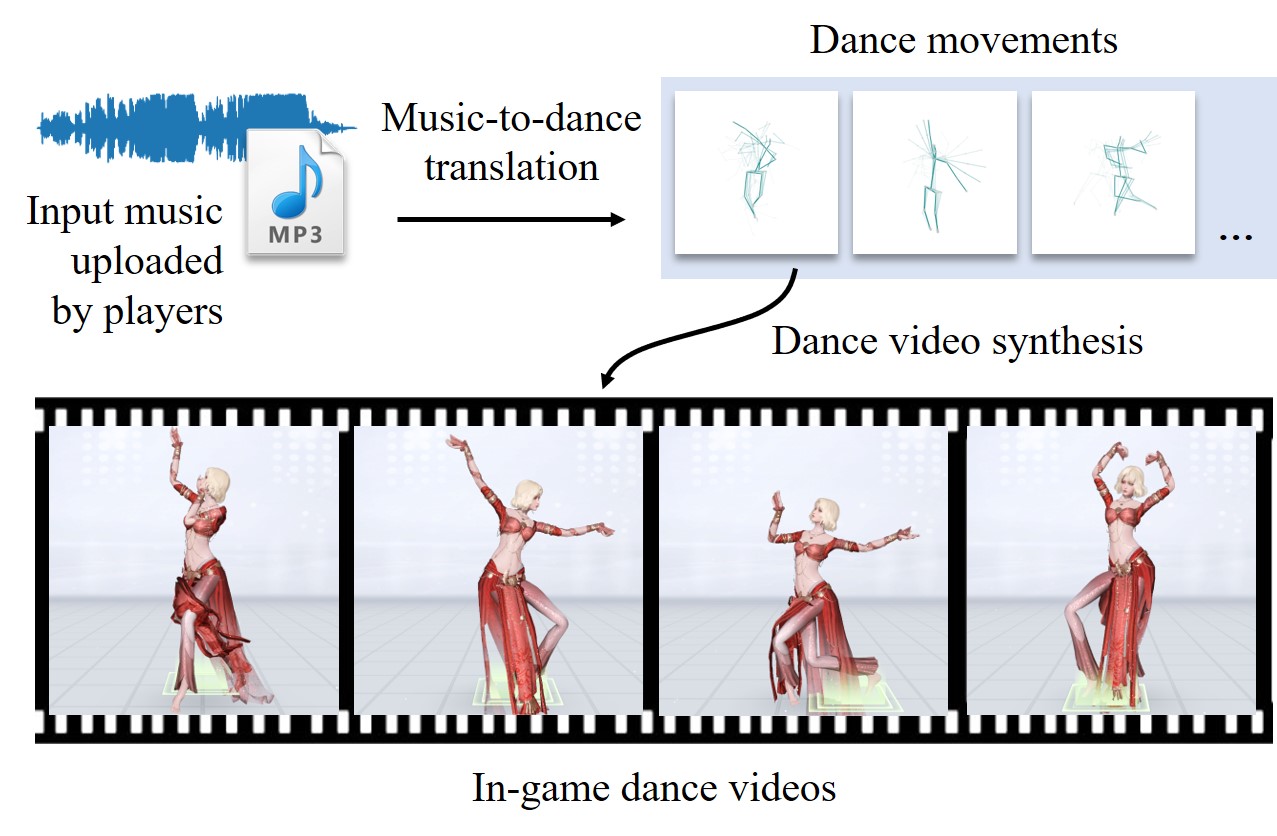}
  \centering
  \caption{An overview of our method: We propose a method for music-to-dance translation based on player uploaded music. We frame the translation as a dance retrieval problem where we firstly segment the music to music phrases and then assign proper dance phrases one by one.}
  \label{fig:teaser}
\end{figure}

The music-dance is a very popular feature for many Role-Playing Games (RPGs), where the players can control their character to dance with the music (e.g. ``Just Dance\footnote{https://www.ubisoft.com/en-us/game/just-dance-2020/}'' and ``FINAL FANTASY XIV\footnote{https://www.finalfantasyxiv.com/}''). Recent games like ``Heaven mobile \footnote{http://tym.163.com/}'' further enriched this feature, where various instruments and pre-defined dance movements are provided. Players can edit vivid music-dance and share it on their social networks. However, the editing and customization of music and dance require a lot of expertise. For those players without experience in such area, choreography for game characters would be a very difficult task. Even for a very experienced team in music-dance, from the early capture of dance movements to the late software synthesis, the entire production time period usually takes several days. In this paper, we investigate an interesting problem called ``Music-to-dance translation'' which aims to automatically generate dance movements for game characters according to the player-uploaded music.  

Recently, music-to-dance translation has drawn increasing research attention due to its wide applications in the game industry and virtual reality. Deep learning based methods have shown great potential in this task~\cite{alemi2017groovenet,tang2018anidance,acmmm20ren}. However, these methods are difficult to apply to in-game expert-level music-to-dance applications. The reason is threefold. First, in choreography theory, dance movements are typically expressed trough the ``strength'', ``speed'' and ``amplitude'' of the human body, while the movements generated by previous methods are mostly based on the amplitude and thus the generation lacks a sense of strength. Second, most previous methods are designed to be trained under a fully-supervised fashion and require a large amount of motion data captured in advance. However, capturing dance motions is usually expensive, time-consuming, and requires the assistance of professional dancers. Finally, previous methods cannot provide players with an interactive experience.

To solve the above problems, we propose a novel method for generating high-quality music-dances. We symbolize the dance movements and re-formulate the music-to-dance translation as a phrase-wise dance phrase retrieval problem. Different with the dance generative models that directly generate the dance movements from the music, we consider the dance movements as a set of semantic fragments according to the choreography theory, and then arrange these phrases for music fragments one by one. To map music phrases to dance moves, we build an encoder-decoder network that takes in the Mel Spectrogram of a music phrase and then predicts the corresponding index of the dance phrase. As a temporal prediction problem, we introduce ``transition priors'' of the dance phrases based on a first-order Markov model to improve the context reasoning, where the transition matrices are used to re-scale the probability of predicted results and get a smoother and more consistent generation result. 

Considering the high cost of building large-scale dance movements datasets, we take advantages of the semi-supervised learning~\cite{oliver2018realistic}, to improve the robustness and generalization ability of our method. We extend our method on a large unlabeled music dataset (20x larger than our labeled one). We first train our method on this unlabeled music dataset with self-supervised pretext tasks. We enforce the network reconstruct the music phrases as well as its melody and rhythm from the latent representations. The model can be thus pre-trained to learn a good representation of the music phrases from the pretext tasks we designed without human annotations. After the pre-training, we fine-tune the model on a labeled subset. Since the transition matrices initially learned on the labeled data are half-baked, we propose a co-ascent mechanism to jointly refine the transition priors of movements and improve the accuracy of the prediction. Specifically, we use the transition matrices to correct the prediction results, i.e. generating pseudo labels~\cite{lee2013pseudo} on the large unlabeled dataset, and then iteratively update the matrices and train our networks based on corrected labels. With the help of semi-supervised learning, our method can better generalize to in-the-wild music data. Such scalability is not considered and supported in previous methods. 

Our contributions are summarized as follows: \par

\begin{itemize}
\item We propose a new music-to-dance translation method based on semi-supervised learning. We extend our method to a larger unlabeled music dataset and explore the effectiveness of self-supervised pre-training in our task. We show that by pre-training the model on the unlabeled dataset and then fine-tune on a labeled subset, the music-to-dance translation accuracy can be greatly improved than that trained solely on the labeled subset from scratch. \par

\item  We introduce a co-ascent mechanism and make full use of the latent structure of the unlabeled data in fine-tuning. We consider the ``transition priors'' of the dance phrases and design a self-correction method to generate pseudo-labels for unlabeled data. To our best knowledge, there are few works that incorporate such a mechanism in this task.\par

\item Different from previous methods where the dance movements are directly generated based on the music, we symbolize the dance movements and re-formulate the music-to-dance translation as a phrase-wise music-to-dance retrieval problem with the guidance of music-dance domain knowledge. With such a design, players can optionally edit the dance moves on top of the generation results according to their preference while such interactivity was ignored in previous methods.
\end{itemize}

\section{Related works}
\subsection{Music-to-dance translation}

Music-to-dance is an emerging research hot-spot in recent years. As a cross-modality generation problem, music-to-dance requires high consistency between music and generated dance on artistic conception. Early works usually adopt statistical models to achieve this goal~\cite{shiratori2006dancing,ofli2008audio,ofli2011learn2dance,fan2011example,lee2013music}. With the development of deep learning, artistic consistency now can be achieved by building supervised deep learning models~\cite{alemi2017groovenet,tang2018dance,NIPS2019dancing}. For example, Alemi \etal first propose GrooveNet to achieve real-time music-driven dance movements generation~\cite{alemi2017groovenet}. In their method, the Factored Conditional Restricted Boltzmann Machines (FCRBM) is reformulated under a Recurrent Neural Network framework and predicts the current motion capture frame by taking in the current music features and the historical frames. Tang \etal further propose an LSTM based Auto-Encoder model named ``Anidance'' to regress motions from acoustic features \cite{tang2018anidance, tang2018dance}. In their method, an extractor is firstly used to reduce the dimension of acoustic features and then a predictor is adopted to translate reduced features to motions. Lee \etal propose a decomposition-to-composition framework for music-to-dance generation~\cite{NIPS2019dancing}, where they use a VAE to model dance units and use a Generative Adversarial Network (GAN) to organize the dance units based on input music. Ren \etal integrate the local temporal discriminator and the global content discriminator for helping generate coherent dance sequences based on the noisy dataset, and then use pose-to-appearance mapping to generate human dance videos~\cite{acmmm20ren}. However, all the above methods directly generate the dance movements from music, which inevitably leads to a problem of motion degradation and is not yet able to meet the requirements of expert-level music-to-dance translation. In this paper, different from previous methods, we symbolize the dance movements and re-formulate the music-to-dance generation as a retrieval problem to avoid the degradation problem. The players can therefore obtain high-quality dance movements arranged by their input music.\par

\begin{figure*}
  \includegraphics[width=0.95\linewidth]{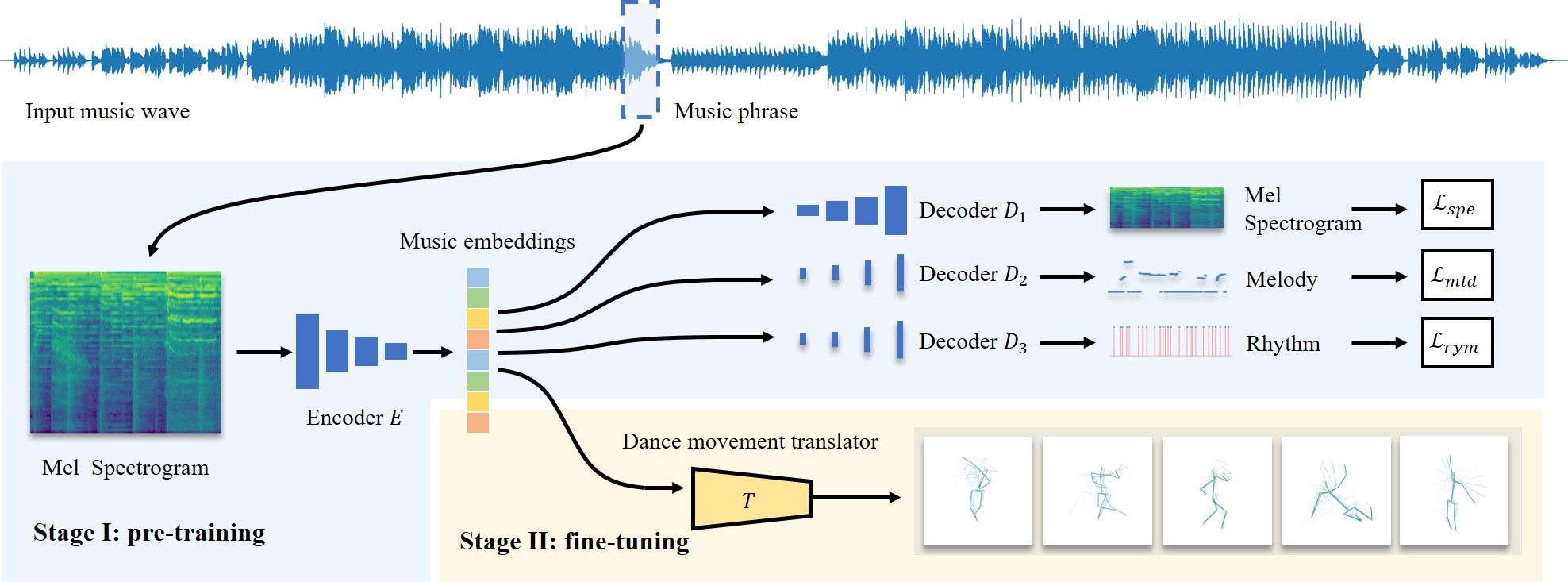}
  \centering
  \caption{An overview of our method. Our method consists of a music encoder $E$ and a dance phrase predictor $T$. We also introduce three decoders for self-supervised pre-training. In the pre-training stage, we train our encoder on a large unlabeled music dataset with three pretext losses - a spectrogram reconstruction loss $\mathcal{L}_{spe}$, a melody prediction loss $\mathcal{L}_{mld}$, and a rhythm prediction loss $\mathcal{L}_{rym}$. In the fine-tuning/inference stage, we train the predictor $T$ on a labeled dance-music dataset so that to translate the input music phrases to dance phrases.}
  \label{fig:pipeline}
\end{figure*}

\subsection{Semi-supervised learning}

Semi-supervised learning forms a challenging but important foundation of machine learning methods~\cite{gammerman2013learning,joachims1999transductive,joachims2003transductive,zhu2003semi,bengio2006label} that combines a small amount of labeled data with a large amount of unlabeled one during training to improve the prediction. In recent years, there are various of methods proposed in this field~\cite{oliver2018realistic}. \emph{Consistency regularization methods} aim at building a low-dimensional manifold for unlabeled data. Such a group of methods include $\Pi$-Model~\cite{laine2016temporal,sajjadi2016regularization}, Mean Teacher~\cite{tarvainen2017mean}, Virtual Adversarial Training~\cite{miyato2018virtual}, and etc. \emph{Entropy-based methods} encourage networks have a higher confident, i.e. low-entropy, on all examples by introducing entropy minimization losses~\cite{grandvalet2005semi,pereyra2017regularizing}. \emph{Pseudo-Labeling} is another simple but widely used strategy in semi-supervised learning, which requires that the model can provide probabilistic results for the unlabeled data and then adopt those pseudo-labels with large enough confidence as targets to further train the model~\cite{lee2013pseudo}. After the era of deep learning, semi-supervised learning was used to solve various computer vision tasks, including image classification~\cite{li2019learning,yalniz2019billion}, semantic segmentation~\cite{papandreou2015weakly,kalluri2019universal}, and object detection~\cite{jeong2019consistency}. Semi-supervised learning was also widely used in various tasks in the multimedia field, such as music analysis~\cite{song2007semi,poria2013music,li2004semi}, image understanding~\cite{li2019learning,papandreou2015weakly}, and etc. In this work, we combine the domain knowledge in music-dance with the idea proposed by Lee \etal, and use pseudo-labels to extend our method on a large unlabeled music dataset.\par

\section{Methodology}

In this paper, we propose a simple but efficient semi-supervised learning method for music-to-dance translation. Fig. \ref{fig:pipeline} shows an overview of our method. Our method consists of a music feature encoder, a dance phrase predictor, and several decoders. The encoder is a ResNet50-based~\cite{he2016deep} convolutional network which is trained to encode the Mel Spectrogram of music phrases into music embeddings. The predictor is an attention based fully connected network which takes in the embeddings and predicts dance phrases. The decoders are specifically designed for the pre-training task and will not be involved during the inference stage. 

Given a piece of music (e.g., a pop song), we first segment the music into several phrases. Then, we pre-train our encoder with self-supervised losses on a large unlabeled music dataset. Then, we fine-tune the predictor on labeled music data to assign dance phrases based on the input features. We further design and incorporate a co-ascent mechanism for making full-use of the unlabeled data and improve the translation.

\subsection{Music phrase segmentation}

In choreography, the music phrase is a segment of the music containing complete semantic-level structure and the dance phrases in each music phrase usually represent similar conceptions. 

\begin{figure}
  \includegraphics[width=0.9\linewidth]{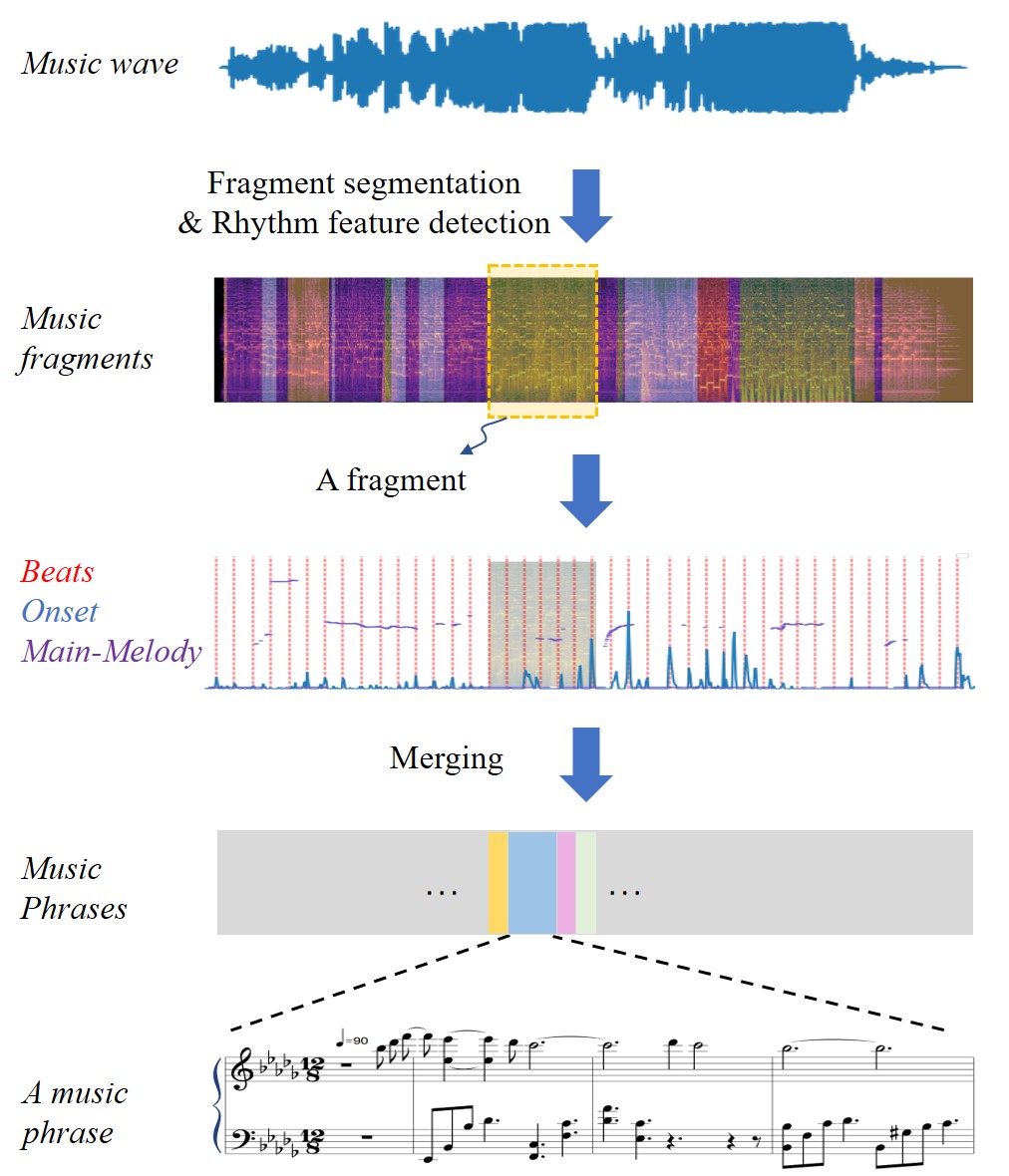}
  \centering
  \caption{The processing pipeline of music phrase segmentation. We firstly segment music to fragments, and then extract features from music fragments. Finally, we slice music phrases based these musical features.}
  \label{fig:music_phrase_seg}
\end{figure}

We thus define the music phrases as our basic processing units in our retrieval model. Considering that there are various types of time signatures for music (e.g., $\frac{2}{4}$, $\frac{3}{4}$, and etc.) and a music phrase may consist of 2$\sim$8 bars (i.e., 6$\sim$24 beats if the time signature is $\frac{3}{4}$), to obtain the segmentation of the music phrases under various beats, we design the following three steps for segmentation, as shown in Fig. \ref{fig:music_phrase_seg}:

\begin{itemize}
\item Long fragment segmentation: Firstly, we analyze the music structure by using spectral clustering and segment music into long fragments. The segmentation on this step is implemented based on librosa~\cite{mcfee2015librosa}.

\item Rhythm feature detection: Secondly, we extract beats and onset by using librosa, and extract main-melody by a deep learning method~\cite{hsieh2019streamlined}.

\item Merging: Finally, we merge the above features and music can be segmented into a set of music phrases - We detect and slice the breaking point of a piece of music judging by melody and onset around a beat.

\end{itemize}

\subsection{Self-supervised pre-training}
\label{sec:pretrain}

The training of our method consists of two stages. In the first stage, we pre-train the encoder on a large unlabeled dataset (music without dance movements) with self-supervised pretext losses. In the second stage, we fix the encoder and fine-tune the predictor on a labeled dataset (music phrases and corresponding dance movements).

Considering that choreography requires the concordance of music-dance on rhythm and melody, we design three pretext tasks for the pre-training - a \textit{spectrogram reconstruction} tasks, a \textit{melody prediction} task, and a \textit{rhythm prediction} task. The pre-training is performed solely on the music data without any human annotations.

{\bf Spectrogram reconstruction}. We compute the Mel Spectrogram for an input music phrase and convert the 1d music signal to a 2D ``image'' by using librosa~\cite{mcfee2015librosa}. We then feed the spectrogram to our ResNet encoder $E$ to produce a set of low dimensional feature embeddings. Because we expect the embeddings containing all information of the input music phrase, we introduce a decoder $D_1$, to upsample the features and restore the spectrogram. We force the Mel Spectrogram before the encoder and after the decoder unchanged. We define the reconstruction loss as follows:
\begin{equation}\label{eq:pretrain-spectrogram}
    \mathcal{L}_{spe}(E, D_1) = {\| D_1(E(\text{Mel}(x))) - \text{Mel}(x)\|}_1,
\end{equation}
where $x$ is the music phrase and $\text{Mel}(x)$ is its Mel Spectrogram. The decoder $D_1$ has a similar structure as the generative network DCGAN~\cite{dcgan}, with 8 transposed 2D-convolution layers.

{\bf Melody prediction}. Main-melody defines the pitch contours of the polyphonic music, and can be used in some high-level tasks such as song identification \cite{serra2010audio}, music genre classification \cite{salamon2012musical}, etc. Different from the previous method~\cite{tang2018anidance} that uses vanilla melody, we use the Main-Melody extracted by deep learning method~\cite{hsieh2019streamlined} to improve the robustness. We define the prediction loss as follows:
\begin{equation}\label{eq:pretrain-melody}
    \mathcal{L}_{mld}(E, D_2) = {\| D_2(E(\text{Mel}(x))) - \text{Melody}(x)\|}_1,
\end{equation}
where $D_2$ is a decoder with 5 transposed 1D-convolution layers for regressing the melody from the embeddings. $\text{Melody}(x)$ is the pre-computed target melody from the music phrase $x$. 

{\bf Rhythm}. We define another prediction head to predict the rhythm from the music embeddings. The prediction loss is defined as follows:
\begin{equation}\label{eq:pretrain-rhythm}
    \mathcal{L}_{rym}(E, D_3) = BCELoss(D_3(E(\text{Mel}(x))), \text{Rythm}(x))
\end{equation}
where BCELoss denotes the Binary-Cross-Entropy-Loss, $D_3$ is a rhythm decoder which has a similar structure as $D_2$ but produces binary output, and $\text{Rythm}(x)$ is the target rhythm from the music phrase $x$, which is pre-computed based on librosa~\cite{mcfee2015librosa} and main-melody. 

{\bf Final pre-training loss} By combining the loss term (\ref{eq:pretrain-spectrogram}), (\ref{eq:pretrain-melody}) and (\ref{eq:pretrain-rhythm}), we define the final pre-training loss as follows:
\begin{equation}
\begin{split}
    &\mathcal{L}_{pre-tr}(E, D_1, D_2, D_3)\\ 
    =& \beta_1\mathcal{L}_{spe} + \beta_2\mathcal{L}_{mld} + \beta_3\mathcal{L}_{rym},
\end{split}
\end{equation}
where $\beta_1$, $\beta_2$, and $\beta_3$ are the weights to balance the loss terms. We train the encoder $E$ and the decoders ($D_1$, $D_2$, $D_3$) to minimize the above loss function. After the pre-training, we remove the decoders and only keep the weights of the encoder for a further fine-tuning on music-dance data pairs.

\subsection{Dance phrase prediction}
\label{sec:dance-predict}

We build an attention-based multilayer perceptron as our dance phrase predictor $T$. The $T$ consists of three residual attention blocks and two Fully Connected (FC) layers. In each of the block, we make a simple modification of the squeeze and excitation block in SENet~\cite{hu2018squeeze} to apply it to an FC layer (the global pooling layer thus is removed).

The $T$ is trained to predicts the index of a proper dance phrase. For each music phrase, we define the prediction loss as the cross-entropy loss between the predicted probability distribution and the $K$ possible dance phrases captured in the dance library:
\begin{equation}\label{eq:fine-tune-loss}
    \mathcal{L}_{pred} = -\sum_{i=1}^K \hat{y}_p^{(i)} \log (F_{pred}(\bm{u})^{(i)}),
\end{equation}
where $[\hat{y}_p^{(1)}, ..., \hat{y}_p^{(K)}]$ represent the one-hot ground truth vector of the prediction. $F_{pred}(\bm{u})^{(i)}$ represents the predicted probability for the $i$th kind of dance phrase. $\bm{u}=E(\text{Mel}(x))$ is the music embedding from the encoder $E$. We train the encoder and predictor from the self-supervised pre-trained initialization. During the training, we fix the encoder $E$ and only update the predictor $T$ for a faster convergence.

\subsection{Co-ascent learning}

Once we have built the above retrieval model, the music-to-dance translation essentially becomes a phrase-wise retrieval problem. Considering that building a large scale dance phrase dataset is very expensive, we introduce the co-ascent learning mechanism to migrate our learning process to unlabeled data. This method also improves the prediction by using context reasoning.

{\bf Transition matrix}. Inspired by the N-gram~\cite{brown1992class} that has been widely used in the field of Natural Language Processing, we introduce a dance phrase transition matrix $\mathbf{M}\in \mathbb{R}^{K\times K}$ to capture the probability transition between the two adjacent dance phrases. This matrix can be seen as having a similar meaning to the probability transition matrix in the first-order Markov process. During the inference stage, we use this matrix to re-scale the prediction results of the current phrase (based on the history predictions). The re-scale of the predicted class probability can be written as follows:
\begin{equation}
\begin{split}
P(d_t| \bm{u}_t, d_{t-1}) & = P(d_t|\bm{u}_t) P(d_t|d_{t-1}) \\
                    & = F_{pred}(\bm{u}_t)\mathbf{M}(d_{t-1} \rightarrow d_t),
\end{split}
\label{eq:transfer}
\end{equation}
where $d_t$ is the dance phrase at the time step $t$, $P(d_t | \bm{u}_t, d_{t-1})$ is the re-scaling results, $F_{pred}(\bm{u}_t)$ is the raw prediction results of the prediction head $F_{pred}$, and $M(d_{t-1} \rightarrow d_t)$ is the transition probability between two dance phrases from the step $t-1$ to $t$.

\begin{figure}
  \includegraphics[width=0.9\linewidth]{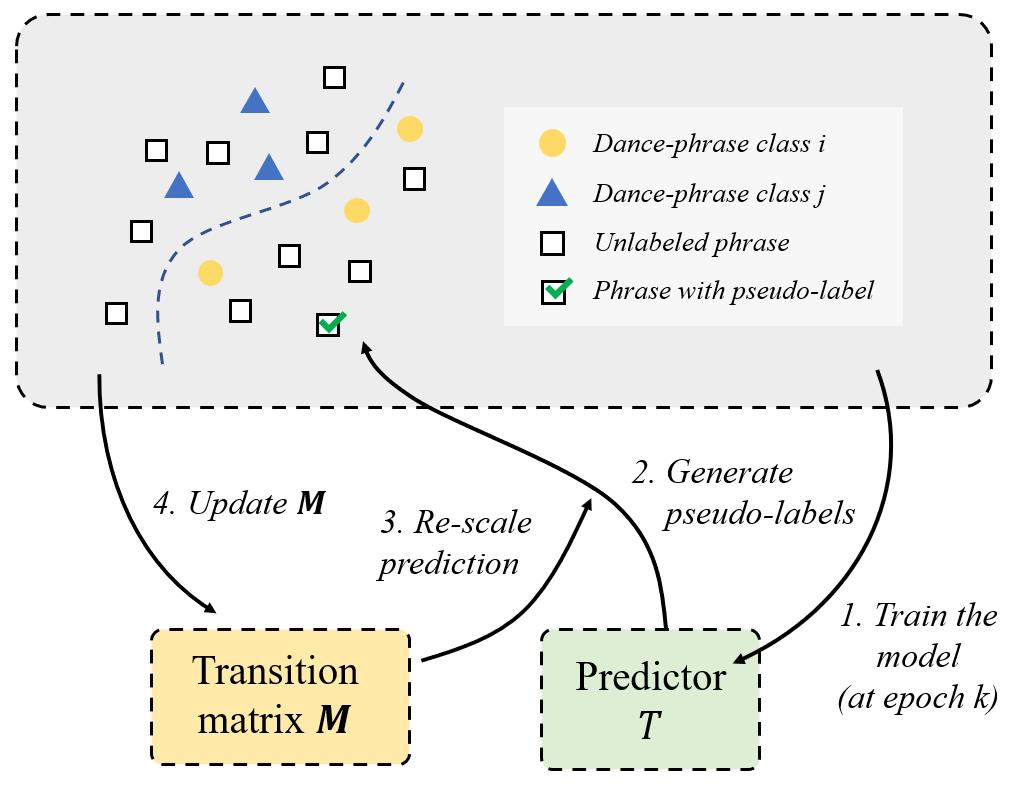}
  \centering
  \caption{The pipeline of the proposed co-ascent learning. We further train our predictor in a semi-supervised manner, where the proposed transition matrix is also integrated to correct the pseudo-labels and also to be jointly updated.}
  \label{fig:Co-ascent-learning}
\end{figure}

{\bf Co-ascent learning}. Pseudo-labeling~\cite{lee2013pseudo} is a simple but effective strategy that has been widely used in semi-supervised learning methods. In our method, we first train the networks on a small labeled dataset and then apply the weak model to all unlabeled data (music without dances) to predict the corresponding labels. The dataset with both true labels and pseudo labels is again used to train the network to enhance the decision boundary. During the pseudo-labeling process, we also apply the transition matrix $\mathbf{M}$ to correct the predictions of our network, and the corrected labels are further used to update the transition matrix. The update of the transfer matrix is performed based on the product of the confidences of two pseudo-labeled music phrases:
\begin{equation}
    M_{k+1}(d_{t-1} \rightarrow d_{t}) = M_k(d_{t-1} \rightarrow d_{t}) + P(d_{t-1}) P(d_{t})
    \label{eq:M-update}
\end{equation}
where $M_{k+1}$ is the transition matrix after $k$th updates by using the pseudo-labels. $P(d_{t})$ is the prediction confidence on the dance phrase at the time step $t$. Since the transition matrix and the networks can be mutually improved based on Eq.~\ref{eq:transfer} and Eq.~\ref{eq:M-update}, we refer to this mechanism as co-ascent learning. 

\begin{figure*}[h]
  \includegraphics[width=0.9\linewidth]{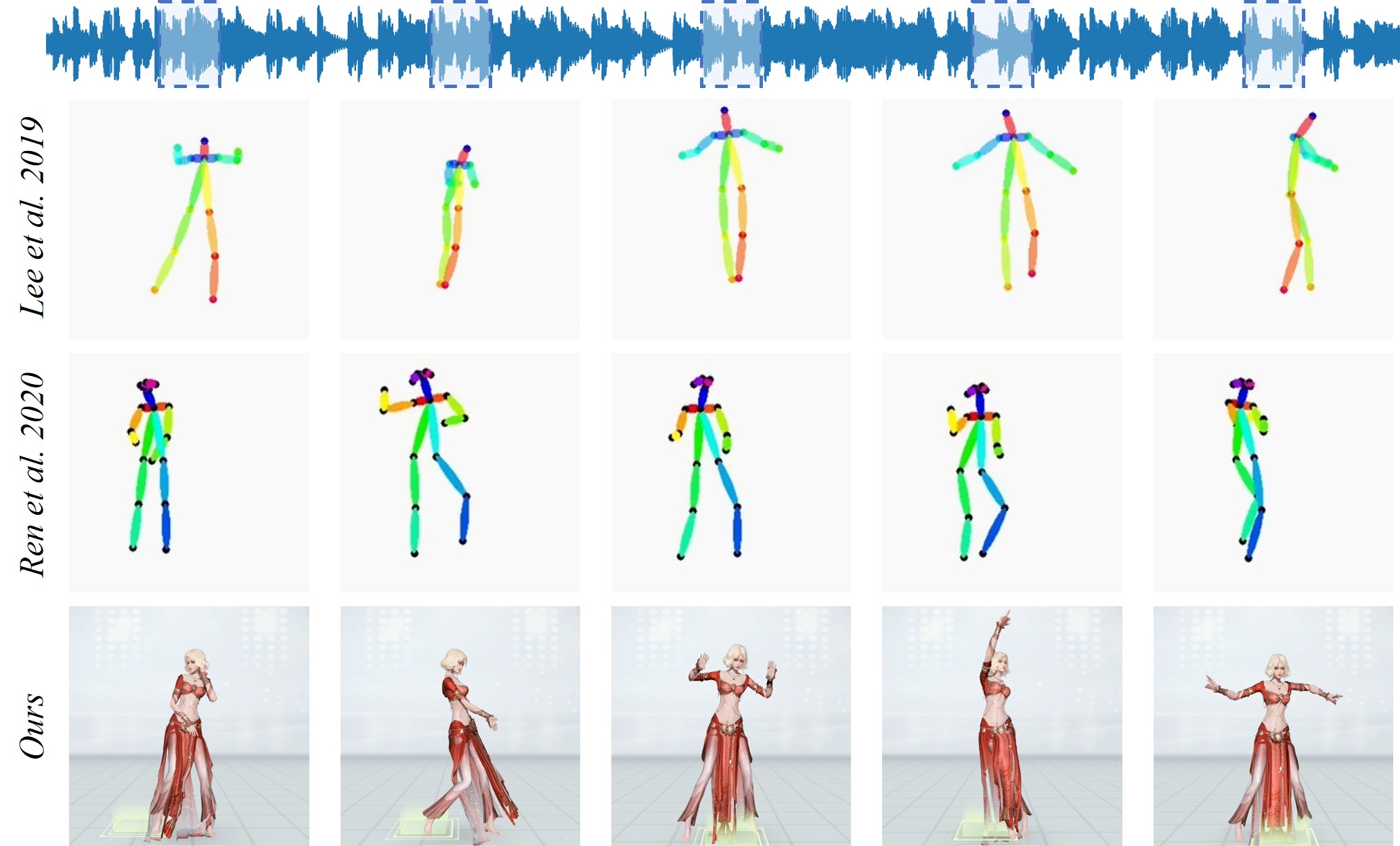}
  \centering
  \caption{Comparisons between our method (shown in game) and previous methods on the music ``Sorry''.}
  \label{fig:rst}
\end{figure*}

\subsection{Implement details}

{\bf Training details.}

In our method, we adopt Mel Spectrogram as the input music feature rather than Mel-frequency cepstral coefficients (MFCCs) because it contains more original music information, and we aim to learn a better representation of music to replace manual features (i.e. MFCCs~\cite{logan2000mel}). The input Mel Spectrogram is resized to $128\times128$ before fed into the encoder $E$, the melody and rhythm are also resized to $1\times128$. The dimension of music embeddings produced by the encoder is set to 512. For a detailed network configuration and the co-ascent learning pipeline, please refer to our Appendix.

In the pre-training stage, we use Adam optimizer~\cite{kingma2014adam} to train our model with the learning rate of $10^{-4}$. The learning rate decay is set to 0.1 per 50 epochs and the training stops at 200 epochs. We set the loss coefficient $\beta_1 = \beta_2 = 1$ and $\beta_3 = 10$. In the supervised fine-tuning stage, we train our translator by SGD with the learning rate of $10^{-2}$, momentum $0.9$, weight decay $5\times10^{-4}$ and the max-epoch number of $500$. In the co-ascent stage, we set the learning rate to $10^{-5}$, update pseudo labels every $5$ epochs, initialize the transition matrix $\mathbf{M}$ based on the style of dance phrases (i.e. the similar dance moves are allowed to transfer) and further clip the range of $\mathbf{M}$ within $[0.01, 1]$ to improve stability. Other configurations are kept the same as our supervised fine-tuning stage.

{\bf Blending of dance phrases}. Considering that the dance moves in adjacent phrases are not always able to connect end to end, we use a common technique called blending\footnote{https://unity.com/} to smooth the movements on switching from one dance move to another.

\begin{table*}[h]
\centering
\caption{The experimental results of the ablation studies (Higher score indicates better performance)}
\begin{tabular}{c|C{2.5cm}C{1.5cm}C{1.5cm}C{2cm}|C{1.5cm}C{1.5cm}C{1.5cm}}
\toprule
\multirow{2}{*}{\textbf{Group}} & \multicolumn{4}{c|}{\textbf{Ablations}} & \multicolumn{3}{c}{\textbf{Index}}\\
& Self-Supervised & Attention & Balance & Co-Ascent & Top1 & Top5 & Top10 \\
\midrule
\uppercase\expandafter{\romannumeral1} &$\times$ & $\times$ & $\times$ & $\times$ & $12.3\%$ & $20.5\%$ & $23.6\%$  \\
\uppercase\expandafter{\romannumeral2} &$\checkmark$ & $\times$ & $\times$ & $\times$ & $14.5\%$ & $19.3\%$ & $22.3\%$  \\
\uppercase\expandafter{\romannumeral3} &$\checkmark$ & $\checkmark$ & $\times$ & $\times$ & $19.1\%$ & $23.7\%$ &  $25.5\%$  \\
\uppercase\expandafter{\romannumeral4} & $\checkmark$ & $\checkmark$ & $\checkmark$ & $\times$ & $\underline{19.3\%}$ & $\mathbf{25.0\%}$ & $\mathbf{27.2\%}$  \\
\uppercase\expandafter{\romannumeral5} &$\checkmark$ & $\checkmark$ & $\checkmark$ & $\checkmark$ & $\mathbf{19.8\%}$ & $\underline{24.8\%}$ &  $\underline{26.8\%}$  \\
\bottomrule
\end{tabular}%
\label{tab:ablation}
\end{table*}%

\begin{table*}[h]
\centering
\caption{The experimental results of the subjective evaluation (Closer to Rank 1 represents better performance)}
\begin{tabular}{c|ccc|c}
\toprule
\multirow{2}{*}{\textbf{Method}} & \multicolumn{3}{c|}{\textbf{Ranking}} & \multirow{2}{*}{\textbf{Average Ranking}}\\
& Group1 & Group2 & Unseen &  \\
\midrule
Dancing to music~\cite{NIPS2019dancing}  & $ 2.94 \pm 0.33 $ & $ 3.00 \pm 0.00 $ & $ 3.00 \pm 0.00 $ & $ 2.98 \pm 0.21 $  \\
Dance Video Synthesis~\cite{acmmm20ren} & $ 1.75 \pm 0.49 $ & $ 1.81 \pm 0.40$ & $ 1.78 \pm 0.42 $ & $ 1.78 \pm 0.44 $  \\
Ours & $ \mathbf{1.31 \pm 0.46} $ & $ \mathbf{1.19 \pm 0.40} $ & $ \mathbf{1.22 \pm 0.42} $ & $ \mathbf{1.24 \pm 0.43} $ \\
\bottomrule
\end{tabular}%
\label{tab:objective}
\end{table*}%

\section{Experiments}

\subsection{Dataset and experimental setup}

We test our method on the music-dance creation platform of a role-playing game named ``Heaven mobile''. We build two datasets for our task: \par

{\bf Labeled Dance-Music Dataset}. 
In this dataset, we first recorded 1,101 different dance phrases by using motion capturing devices (Vicon V16 cameras). Five professional dancers took part in the motion capture for one month. We then collected about 600 songs ($\sim$33 hours) with different genres that are suitable for choreography. We segment these songs into about 16773 music phrases and invite six experts to arrange dance phrases for these music phrases song by song (multiple different music phrases may correspond to the same kind of dance phrases). For performance evaluation, we split this dataset into a training set (90 \%) and a test set (10 \%). \par

{\bf Unlabeled Music Dataset}. 
In addition to the labeled dataset, we also collected an unlabeled dataset which is 20x larger than the labeled one. The dataset consists of about 10k songs in various styles ($\sim$686 hours). We segment each song of this dataset into music phrases and finally 293,579 music phrases are extracted and orderly packaged.

\subsection{Music-to-dance translation results}

Fig.~\ref{fig:rst} shows a group of translation results by using our method and previous state of the art methods on the music ``Sorry'' (also used in the previous work~\cite{acmmm20ren}). It can be seen that the music-dance video generated by our method not only accurately capture the rhythm in the song, but also contain rich musical feelings and movement strength.

\subsection{Ablation studies}

The ablation experiments are conducted to verify the importance of each component in our network. We evaluate five configurations of our method, including:

Group I: A ResNet-50 encoder is only adopted and initialized by ImageNet pre-trained weights.

Group II: A ResNet-50 encoder is adopted and initialized by the weights trained under self-supervised learning.

Group III: We fix the encoder trained by self-supervised losses and fine-tuning the attention-based predictor on the labeled dataset.

Group IV: We further balance the labeled dataset on top of Group III.

Group V: We apply co-ascent learning on top of Group IV.

The results are listed in Fig.~\ref{tab:ablation}. We can see that our full implementation (Group V) achieves significant improvement than baselines, the self-supervised learning (Group III) shows a noticeable impact on our results (+6.8\% on top1 than Group I), and only using self-supervised pre-trained weights may lead the overfitting on the small dataset (+2.2\% on top1 than Group I). Besides, co-ascent learning also shows improvements on top1 (+0.5\%) - although the scores are somewhat incremental, we find that co-ascent learning provides prediction results with a much more consistency style.

\subsection{Subjective evaluation}

Since the predictor faces to a 1000-classification problem and the choreography can be very flexible, dance phrases can often exchangeable. In other words, a higher index accuracy in this task may not necessarily indicate better performance (even may indicate overfitting on the proxy task).

To better evaluate the quality of the generated dance phrases, subjective evaluations are further conducted. In this experiment, we first collect three groups of music: 1) music used in the previous method~\cite{acmmm20ren}, 2) music from our unlabeled test set, 3) unseen style music outside of our dataset. Note that all these musics are not shown in our training dataset. Then we generate dance videos based on three methods, i.e. our full implementation method and two previous state of the art methods~\cite{acmmm20ren,NIPS2019dancing}. 

For each group of the result, we invite nine certified dance teachers (with more than 10 years dancing experiences) and nine professional dancers (with $5\sim10$ years experiences) to rank the results of the three methods. The result videos are randomly segmented to a set of 30s clips. The experts were asked to ignore the differences in the appearance of character models and focus on the concordance of music-dance and the continuity of dance phrases. The statistics of the rating for different video groups are listed in Table~\ref{tab:objective}. The experts agree our method generates expert-level dance videos on the fluency and strength of the dance movements. The superiority of our method is twofold: 1) previous methods focus more on generating short sequences ($<$5s) while choreography requires the long-term matching between music and dance phrases, 2) the generated dance movements suffer from a degradation problem, while we void this problem by rethinking this task as a dance phrase retrieval problem which also keeps better interactivity and is more suitable for game applications.

\section{Limitation}

Although we achieve noticeable improvement than previous methods, our method still has limitations. The first limitation is that since the encoder takes in resized square inputs, it drops absolute rhythmic information and may lead to a failure on very smooth music. The second limitation is that since the blending method used in this work is linear, the transition between two dance phrases may cause model clipping on large movement changes. We will focus on these problems in our future work. 

\section{Conclusion}

In this paper, we propose a new method for automatic music-to-dance translation. We re-formulate the music-to-dance translation as a semi-supervised dance movement retrieval problem based on the choreography theory. We also build a new music-dance dataset which consists of over 16k music phrases labeled with dance movements and also 300k unlabeled ones. We design a self-supervised pre-training method and a co-ascent learning pipeline so that to fully explore the information in the unlabeled music data. Our experimental results in our dataset suggest that our methods can generate expert-level music-dances. The ablation studies also suggest the effectiveness of the core design in our method.

{\fontsize{9.0pt}{10.0pt} \selectfont
\bibliography{music2dance}

\begin{thebibliography}{42}
\providecommand{\natexlab}[1]{#1}
\providecommand{\url}[1]{\texttt{#1}}
\providecommand{\urlprefix}{URL }
\expandafter\ifx\csname urlstyle\endcsname\relax
  \providecommand{\doi}[1]{doi:\discretionary{}{}{}#1}\else
  \providecommand{\doi}{doi:\discretionary{}{}{}\begingroup
  \urlstyle{rm}\Url}\fi

\bibitem[{Alemi, Fran{\c{c}}oise, and Pasquier(2017)}]{alemi2017groovenet}
Alemi, O.; Fran{\c{c}}oise, J.; and Pasquier, P. 2017.
\newblock GrooveNet: Real-time music-driven dance movement generation using
  artificial neural networks.
\newblock \emph{networks} 8(17): 26.

\bibitem[{Bengio, Delalleau, and Le~Roux(2006)}]{bengio2006label}
Bengio, Y.; Delalleau, O.; and Le~Roux, N. 2006.
\newblock Label Propagation and Quadratic Criterion, chapter 11.

\bibitem[{Brown et~al.(1992)Brown, Desouza, Mercer, Pietra, and
  Lai}]{brown1992class}
Brown, P.~F.; Desouza, P.~V.; Mercer, R.~L.; Pietra, V. J.~D.; and Lai, J.~C.
  1992.
\newblock Class-based n-gram models of natural language.
\newblock \emph{Computational linguistics} 18(4): 467--479.

\bibitem[{Fan, Xu, and Geng(2011)}]{fan2011example}
Fan, R.; Xu, S.; and Geng, W. 2011.
\newblock Example-based automatic music-driven conventional dance motion
  synthesis.
\newblock \emph{IEEE transactions on visualization and computer graphics}
  18(3): 501--515.

\bibitem[{Gammerman, Vovk, and Vapnik(2013)}]{gammerman2013learning}
Gammerman, A.; Vovk, V.; and Vapnik, V. 2013.
\newblock Learning by transduction.
\newblock \emph{arXiv preprint arXiv:1301.7375} .

\bibitem[{Grandvalet and Bengio(2005)}]{grandvalet2005semi}
Grandvalet, Y.; and Bengio, Y. 2005.
\newblock Semi-supervised learning by entropy minimization.
\newblock In \emph{Advances in neural information processing systems},
  529--536.

\bibitem[{He et~al.(2016)He, Zhang, Ren, and Sun}]{he2016deep}
He, K.; Zhang, X.; Ren, S.; and Sun, J. 2016.
\newblock Deep residual learning for image recognition.
\newblock In \emph{Proceedings of the IEEE conference on computer vision and
  pattern recognition}, 770--778.

\bibitem[{Hsieh, Su, and Yang(2019)}]{hsieh2019streamlined}
Hsieh, T.-H.; Su, L.; and Yang, Y.-H. 2019.
\newblock A streamlined encoder/decoder architecture for melody extraction.
\newblock In \emph{ICASSP 2019-2019 IEEE International Conference on Acoustics,
  Speech and Signal Processing (ICASSP)}, 156--160. IEEE.

\bibitem[{Hu, Shen, and Sun(2018)}]{hu2018squeeze}
Hu, J.; Shen, L.; and Sun, G. 2018.
\newblock Squeeze-and-excitation networks.
\newblock In \emph{Proceedings of the IEEE Conference on Computer Vision and
  Pattern Recognition}, 7132--7141.

\bibitem[{Jeong et~al.(2019)Jeong, Lee, Kim, and Kwak}]{jeong2019consistency}
Jeong, J.; Lee, S.; Kim, J.; and Kwak, N. 2019.
\newblock Consistency-based Semi-supervised Learning for Object detection.
\newblock In \emph{Advances in Neural Information Processing Systems},
  10758--10767.

\bibitem[{Joachims(1999)}]{joachims1999transductive}
Joachims, T. 1999.
\newblock Transductive inference for text classification using support vector
  machines.
\newblock In \emph{Icml}, volume~99, 200--209.

\bibitem[{Joachims(2003)}]{joachims2003transductive}
Joachims, T. 2003.
\newblock Transductive learning via spectral graph partitioning.
\newblock In \emph{Proceedings of the 20th International Conference on Machine
  Learning (ICML-03)}, 290--297.

\bibitem[{Kalluri et~al.(2019)Kalluri, Varma, Chandraker, and
  Jawahar}]{kalluri2019universal}
Kalluri, T.; Varma, G.; Chandraker, M.; and Jawahar, C. 2019.
\newblock Universal semi-supervised semantic segmentation.
\newblock In \emph{Proceedings of the IEEE International Conference on Computer
  Vision}, 5259--5270.

\bibitem[{Kingma and Ba(2014)}]{kingma2014adam}
Kingma, D.~P.; and Ba, J. 2014.
\newblock Adam: A method for stochastic optimization.
\newblock \emph{arXiv preprint arXiv:1412.6980} .

\bibitem[{Laine and Aila(2016)}]{laine2016temporal}
Laine, S.; and Aila, T. 2016.
\newblock Temporal ensembling for semi-supervised learning.
\newblock \emph{arXiv preprint arXiv:1610.02242} .

\bibitem[{Lee(2013)}]{lee2013pseudo}
Lee, D.-H. 2013.
\newblock Pseudo-label: The simple and efficient semi-supervised learning
  method for deep neural networks.
\newblock In \emph{Workshop on challenges in representation learning, ICML},
  volume~3, 2.

\bibitem[{Lee et~al.(2019)Lee, Yang, Liu, Wang, Lu, Yang, and
  Kautz}]{NIPS2019dancing}
Lee, H.-Y.; Yang, X.; Liu, M.-Y.; Wang, T.-C.; Lu, Y.-D.; Yang, M.-H.; and
  Kautz, J. 2019.
\newblock Dancing to music.
\newblock In \emph{Advances in Neural Information Processing Systems},
  3586--3596.

\bibitem[{Lee, Lee, and Park(2013)}]{lee2013music}
Lee, M.; Lee, K.; and Park, J. 2013.
\newblock Music similarity-based approach to generating dance motion sequence.
\newblock \emph{Multimedia tools and applications} 62(3): 895--912.

\bibitem[{Li and Ogihara(2004)}]{li2004semi}
Li, T.; and Ogihara, M. 2004.
\newblock Semi-supervised learning for music artists style identification.
\newblock In \emph{Proceedings of the thirteenth ACM international conference
  on Information and knowledge management}, 152--153.

\bibitem[{Li et~al.(2019)Li, Sun, Liu, Zhou, Zheng, Chua, and
  Schiele}]{li2019learning}
Li, X.; Sun, Q.; Liu, Y.; Zhou, Q.; Zheng, S.; Chua, T.-S.; and Schiele, B.
  2019.
\newblock Learning to self-train for semi-supervised few-shot classification.
\newblock In \emph{Advances in Neural Information Processing Systems},
  10276--10286.

\bibitem[{Logan et~al.(2000)}]{logan2000mel}
Logan, B.; et~al. 2000.
\newblock Mel frequency cepstral coefficients for music modeling.
\newblock In \emph{Ismir}, volume 270, 1--11.

\bibitem[{McFee et~al.(2015)McFee, Raffel, Liang, Ellis, McVicar, Battenberg,
  and Nieto}]{mcfee2015librosa}
McFee, B.; Raffel, C.; Liang, D.; Ellis, D.~P.; McVicar, M.; Battenberg, E.;
  and Nieto, O. 2015.
\newblock librosa: Audio and music signal analysis in python.
\newblock In \emph{Proceedings of the 14th python in science conference},
  volume~8.

\bibitem[{Miyato et~al.(2018)Miyato, Maeda, Koyama, and
  Ishii}]{miyato2018virtual}
Miyato, T.; Maeda, S.-i.; Koyama, M.; and Ishii, S. 2018.
\newblock Virtual adversarial training: a regularization method for supervised
  and semi-supervised learning.
\newblock \emph{IEEE transactions on pattern analysis and machine intelligence}
  41(8): 1979--1993.

\bibitem[{Ofli et~al.(2008)Ofli, Demir, Yemez, Erzin, Tekalp, Balc{\i},
  K{\i}zo{\u{g}}lu, Akarun, Canton-Ferrer, Tilmanne et~al.}]{ofli2008audio}
Ofli, F.; Demir, Y.; Yemez, Y.; Erzin, E.; Tekalp, A.~M.; Balc{\i}, K.;
  K{\i}zo{\u{g}}lu, {\.I}.; Akarun, L.; Canton-Ferrer, C.; Tilmanne, J.; et~al.
  2008.
\newblock An audio-driven dancing avatar.
\newblock \emph{Journal on Multimodal User Interfaces} 2(2): 93--103.

\bibitem[{Ofli et~al.(2011)Ofli, Erzin, Yemez, and
  Tekalp}]{ofli2011learn2dance}
Ofli, F.; Erzin, E.; Yemez, Y.; and Tekalp, A.~M. 2011.
\newblock Learn2dance: Learning statistical music-to-dance mappings for
  choreography synthesis.
\newblock \emph{IEEE Transactions on Multimedia} 14(3): 747--759.

\bibitem[{Oliver et~al.(2018)Oliver, Odena, Raffel, Cubuk, and
  Goodfellow}]{oliver2018realistic}
Oliver, A.; Odena, A.; Raffel, C.~A.; Cubuk, E.~D.; and Goodfellow, I. 2018.
\newblock Realistic evaluation of deep semi-supervised learning algorithms.
\newblock In \emph{Advances in Neural Information Processing Systems},
  3235--3246.

\bibitem[{Papandreou et~al.(2015)Papandreou, Chen, Murphy, and
  Yuille}]{papandreou2015weakly}
Papandreou, G.; Chen, L.-C.; Murphy, K.~P.; and Yuille, A.~L. 2015.
\newblock Weakly-and semi-supervised learning of a deep convolutional network
  for semantic image segmentation.
\newblock In \emph{Proceedings of the IEEE international conference on computer
  vision}, 1742--1750.

\bibitem[{Paszke et~al.(2019)Paszke, Gross, Massa, Lerer, Bradbury, Chanan,
  Killeen, Lin, Gimelshein, Antiga, Desmaison, Kopf, Yang, DeVito, Raison,
  Tejani, Chilamkurthy, Steiner, Fang, Bai, and
  Chintala}]{PYTORCH_NEURIPS2019_9015}
Paszke, A.; Gross, S.; Massa, F.; Lerer, A.; Bradbury, J.; Chanan, G.; Killeen,
  T.; Lin, Z.; Gimelshein, N.; Antiga, L.; Desmaison, A.; Kopf, A.; Yang, E.;
  DeVito, Z.; Raison, M.; Tejani, A.; Chilamkurthy, S.; Steiner, B.; Fang, L.;
  Bai, J.; and Chintala, S. 2019.
\newblock PyTorch: An Imperative Style, High-Performance Deep Learning Library.
\newblock In \emph{Advances in Neural Information Processing Systems 32},
  8024--8035. Curran Associates, Inc.
\newblock
  \urlprefix\url{http://papers.neurips.cc/paper/9015-pytorch-an-imperative-style-high-performance-deep-learning-library.pdf}.

\bibitem[{Pereyra et~al.(2017)Pereyra, Tucker, Chorowski, Kaiser, and
  Hinton}]{pereyra2017regularizing}
Pereyra, G.; Tucker, G.; Chorowski, J.; Kaiser, {\L}.; and Hinton, G. 2017.
\newblock Regularizing neural networks by penalizing confident output
  distributions.
\newblock \emph{arXiv preprint arXiv:1701.06548} .

\bibitem[{Poria et~al.(2013)Poria, Gelbukh, Hussain, Bandyopadhyay, and
  Howard}]{poria2013music}
Poria, S.; Gelbukh, A.; Hussain, A.; Bandyopadhyay, S.; and Howard, N. 2013.
\newblock Music genre classification: A semi-supervised approach.
\newblock In \emph{Mexican Conference on Pattern Recognition}, 254--263.
  Springer.

\bibitem[{Radford, Metz, and Chintala(2015)}]{dcgan}
Radford, A.; Metz, L.; and Chintala, S. 2015.
\newblock Unsupervised representation learning with deep convolutional
  generative adversarial networks.
\newblock \emph{arXiv preprint arXiv:1511.06434} .

\bibitem[{Ren et~al.(2020)Ren, Li, Huang, and Chen}]{acmmm20ren}
Ren, X.; Li, H.; Huang, Z.; and Chen, Q. 2020.
\newblock Self-supervised Dance Video Synthesis Conditioned on Music.

\bibitem[{Sajjadi, Javanmardi, and Tasdizen(2016)}]{sajjadi2016regularization}
Sajjadi, M.; Javanmardi, M.; and Tasdizen, T. 2016.
\newblock Regularization with stochastic transformations and perturbations for
  deep semi-supervised learning.
\newblock In \emph{Advances in neural information processing systems},
  1163--1171.

\bibitem[{Salamon, Rocha, and G{\'o}mez(2012)}]{salamon2012musical}
Salamon, J.; Rocha, B.; and G{\'o}mez, E. 2012.
\newblock Musical genre classification using melody features extracted from
  polyphonic music signals.
\newblock In \emph{2012 ieee international conference on acoustics, speech and
  signal processing (icassp)}, 81--84. IEEE.

\bibitem[{Serra, G{\'o}mez, and Herrera(2010)}]{serra2010audio}
Serra, J.; G{\'o}mez, E.; and Herrera, P. 2010.
\newblock Audio cover song identification and similarity: background,
  approaches, evaluation, and beyond.
\newblock In \emph{Advances in Music Information Retrieval}, 307--332.
  Springer.

\bibitem[{Shiratori, Nakazawa, and Ikeuchi(2006)}]{shiratori2006dancing}
Shiratori, T.; Nakazawa, A.; and Ikeuchi, K. 2006.
\newblock Dancing-to-music character animation.
\newblock In \emph{Computer Graphics Forum}, volume~25, 449--458. Wiley Online
  Library.

\bibitem[{Song, Zhang, and Xiang(2007)}]{song2007semi}
Song, Y.; Zhang, C.; and Xiang, S. 2007.
\newblock Semi-supervised music genre classification.
\newblock In \emph{2007 IEEE International Conference on Acoustics, Speech and
  Signal Processing-ICASSP'07}, volume~2, II--729. IEEE.

\bibitem[{Tang, Jia, and Mao(2018)}]{tang2018dance}
Tang, T.; Jia, J.; and Mao, H. 2018.
\newblock Dance with melody: An lstm-autoencoder approach to music-oriented
  dance synthesis.
\newblock In \emph{Proceedings of the 26th ACM international conference on
  Multimedia}, 1598--1606.

\bibitem[{Tang, Mao, and Jia(2018)}]{tang2018anidance}
Tang, T.; Mao, H.; and Jia, J. 2018.
\newblock AniDance: Real-Time Dance Motion Synthesize to the Song.
\newblock In \emph{Proceedings of the 26th ACM international conference on
  Multimedia}, 1237--1239.

\bibitem[{Tarvainen and Valpola(2017)}]{tarvainen2017mean}
Tarvainen, A.; and Valpola, H. 2017.
\newblock Mean teachers are better role models: Weight-averaged consistency
  targets improve semi-supervised deep learning results.
\newblock In \emph{Advances in neural information processing systems},
  1195--1204.

\bibitem[{Yalniz et~al.(2019)Yalniz, J{\'e}gou, Chen, Paluri, and
  Mahajan}]{yalniz2019billion}
Yalniz, I.~Z.; J{\'e}gou, H.; Chen, K.; Paluri, M.; and Mahajan, D. 2019.
\newblock Billion-scale semi-supervised learning for image classification.
\newblock \emph{arXiv preprint arXiv:1905.00546} .

\bibitem[{Zhu, Ghahramani, and Lafferty(2003)}]{zhu2003semi}
Zhu, X.; Ghahramani, Z.; and Lafferty, J.~D. 2003.
\newblock Semi-supervised learning using gaussian fields and harmonic
  functions.
\newblock In \emph{Proceedings of the 20th International conference on Machine
  learning (ICML-03)}, 912--919.

\end{thebibliography}
}

\newpage
\appendix
\onecolumn
\section{Appendix}

\subsection{Details of network configuration}
In this section, we list the configurations of all networks mentioned in our main paper, i.e. the encoder $E$, the predictor $P$, the 2D-decoder $D_1$ and two 1D-decoders $D_2$ \& $D_3$. Our networks are implemented under PyTorch deep learning framework~\cite{PYTORCH_NEURIPS2019_9015}.

\subsubsection{The configuration of Encoder $E$}

A detailed configuration of ResNet50-based encoder $E$ is listed in Table \ref{table:Encoder-E}. The input size of the Encoder $E$ is $128\times128$ pixels, where the Mel Spectrogram is therefore resized on the time dimension. The outputs of $E$ contain a temporal feature $\bm{f_t} \in \mathbb{R}^{512\times4\times1}$ and an embedding $\bm{u} \in \mathbb{R}^{512\times1\times1}$ from \emph{feature layer} and \emph{embedding layer}.

Specifically, in a $c\times w \times w / s$ Convolution / Deconvolution layer, $c$ denotes the number of filters, $w \times w$ denotes the filter's size and $s$ denotes the filter's stride. In a  $w\times w / s$ Maxpool layer, $w$ denotes the pooling window size, and $s$ denotes the pooling stride. In an $n/s$ Bottleneck block \cite{he2016deep}, $n$ denotes the number of planes, and $s$ denotes the block's stride. In an $(h, w)$ AdaptiveAvgPool2d layer, $h$ and $w$ denote the output dimension of height and width, and ``None'' means the size will be the same as the input. \par

\begin{table}[h]
\centering
\renewcommand\arraystretch{1.2}
\begin{tabular}{l|lccc}
    \toprule    
     & \textbf{Layer} & \textbf{Component} & \textbf{Configuration} & \textbf{Feature Size}\\
    \midrule
    \multirow{9}{*}{\rotatebox[origin=c]{90}{\textbf{Encoder} $E$}}
     & Conv\_1 & Conv2d + BN2d + ReLU & 64x7x7 / 2 & 64x64\\
     & MaxPool & MaxPool & 3x3 / 2 & 32x32\\
     & Conv\_2 & 3 x Bottleneck & 64 / 1 & 32x32\\
     & Conv\_3 & 4 x Bottleneck & 128 / 2 & 16x16\\
     & Conv\_4 & 6 x Bottleneck & 256 / 2 & 8x8\\
     & Conv\_5 & 3 x Bottleneck & 512 / 2 & 4x4\\
     & Conv\_6 & Conv2d & 2048x1x1 / 1 & 4x4\\
     & feature & AdaptiveAvgPool2d & (None, 1) & 4x1\\
     & embedding & AdaptiveAvgPool2d & (1, None) & 1x1\\
    \bottomrule
\end{tabular}
    \caption{A detailed configuration of the Encoder $E$.}
    \label{table:Encoder-E}
\end{table}

\newpage 

\subsubsection{The configuration of Decoder $D_1$}

A detailed configuration of Decoder $D_1$ is listed in Table \ref{table:Decoder-D1}. The input of $D_1$ is the embedding $\bm{u}$ with the length 512, and the output is reconstructed Mel Spectrogram with the size of $128\times128$ pixels.\par

\begin{table}[h]
\centering
\renewcommand\arraystretch{1.2}
\begin{tabular}{l|lccc}
    \toprule    
     & \textbf{Layer} & \textbf{Component} & \textbf{Configuration} & \textbf{Feature Size}\\
    \midrule
    \multirow{8}{*}{\rotatebox[origin=c]{90}{\textbf{Decoder} $D_1$}}
     & Layer\_1 & ConvTranspose2d + BN2d + ReLU & 512x4x4 / 1 & 4x4    \\
     & Layer\_2 & ConvTranspose2d + BN2d + ReLU & 512x4x4 / 2 & 8x8    \\
     & Layer\_3 & ConvTranspose2d + BN2d + ReLU & 256x4x4 / 2 & 16x16  \\
     & Layer\_4 & ConvTranspose2d + BN2d + ReLU & 256x4x4 / 2 & 32x32  \\
     & Layer\_5 & ConvTranspose2d + BN2d + ReLU & 128x3x3 / 1 & 32x32  \\
     & Layer\_6 & ConvTranspose2d + BN2d + ReLU & 128x4x4 / 2 & 64x64  \\
     & Layer\_7 & ConvTranspose2d + BN2d + ReLU & 64x3x3  / 1 & 64x64  \\
     & Layer\_8 & ConvTranspose2d               & 1x4x4   / 2 & 128x128\\
    \bottomrule
\end{tabular}
    \caption{A detailed configuration of the Decoder $D_1$.}
    \label{table:Decoder-D1}
\end{table}

\subsubsection{The configuration of Decoder $D_2$ and $D_3$}

Detailed configurations of the Decoder $D_2$ and $D_3$ are listed in Table \ref{table:Decoder-D1}. The input of $D_2$ and $D_3$ is the temporal feature $\bm{f_t}$, and the output is the reconstructed Main-Melody and Rhythm with the length $128$. Since rhythm prediction can be considered as a binary classification problem, we further add a sigmoid function at the end of the Decoder in this task. Similar to the above tables, in a $c\times w / s$ of 1D-Convolution / 1D-Deconvolution layer, $c$ denotes the number of filters, $w$ denotes the filter's length and $s$ denotes the filter's stride.\par

\begin{table}[h]
\centering
\renewcommand\arraystretch{1.2}
\begin{tabular}{l|lccc}
    \toprule    
     & \textbf{Layer} & \textbf{Component} & \textbf{Configuration} & \textbf{Feature Length}\\
    \midrule
    \multirow{6}{*}{\rotatebox[origin=c]{90}{\textbf{Decoder} $D_2$ \& $D_3$}}
     & Layer\_1 & ConvTranspose1d + ReLU & 512x2 / 2 & 8\\
     & Layer\_2 & ConvTranspose1d + ReLU & 256x2 / 2 & 16\\
     & Layer\_3 & ConvTranspose1d + ReLU & 128x2 / 2 & 32\\
     & Layer\_4 & ConvTranspose1d + ReLU & 64x2 / 2 & 64\\
     & Layer\_5 & ConvTranspose1d + ReLU & 32x2 / 2 & 128\\
     & Output   & Conv1d                 & 1x1 / 1 & 128\\
    \bottomrule
\end{tabular}
    \caption{A detailed configuration of the Decoders $D_2$ and $D_3$.}
    \label{table:Decoder-D23}
\end{table}

\newpage

\subsubsection{The configuration of Predictor $T$}

\begin{figure}[h]
    \centering{\includegraphics[width=0.5\linewidth]{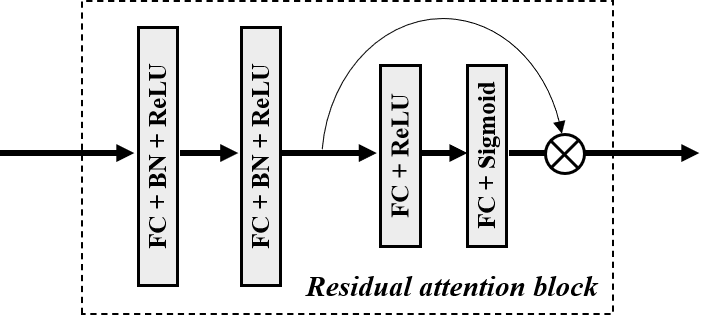}} \\
    \caption{The details of residual attention blocks (Res-Att).}
    \label{fig:res-att}
\end{figure}

In our predictor, we adopt three residual attention blocks and two fully connected layers. The detailed configuration is shown in Table~\ref{table:Predictor-T}, the $(n, m)$ of a Linear and a ``Res-Att'' layer represents that the input and output channel number are $n$ and $m$ respectively, and $K$ is the number of output dance phrases. We follow the  ResNet \cite{he2016deep} and SENet \cite{hu2018squeeze} and set the four fully connected layers in our residual attention blocks (as shown in Fig.~\ref{fig:res-att}) are orderly set to ``Linear(512,1024)'', ``Linear(1024,512)'', ``Linear(512,16)'' and ``Linear(16,512)''.\par

\begin{table}[h]
\centering
\renewcommand\arraystretch{1.2}
\begin{tabular}{l|lccc}
    \toprule    
     & \textbf{Layer} & \textbf{Component} & \textbf{Configuration} & \textbf{Feature Channel}\\
    \midrule
    \multirow{5}{*}{\rotatebox[origin=c]{90}{\textbf{Predictor} $T$}}
     & Layer\_1 & Linear & (512, 512) & 512\\
     & Layer\_2 & Res-Att & (512, 512) & 512\\
     & Layer\_3 & Res-Att & (512, 512) & 512\\
     & Layer\_4 & Res-Att & (512, 512) & 512\\
     & Output   & Linear & (512, K) & K\\
    \bottomrule
\end{tabular}
    \caption{A detailed configuration of the Predictor $T$.}
    \label{table:Predictor-T}
\end{table}

\newpage 

\subsection{Details of co-ascent learning}

In this section, we give a detailed description on our co-ascent learning method, which can notably improve the performance of our music-to-dance translation. The algorithm flow of co-ascent learning is shown in Alg.~\ref{alg:co-ascent}.

\begin{algorithm}[h]
 \KwData{Labeled dataset $D_l$ with $K$ kinds of dance phrases, Unlabeled dataset $D_u$.}
\BlankLine
{\bf Init:} Fix the Encoder $E$ and initialize the predictor $T$ by training $T$ on $D_l$. Calculate the transition matrix $\mathbf{M}_0$ on $D_l$ based on the style of dance phrases. Set the threshold $\tau=0.9$ and momentum parameter $\alpha=0.5$\;
\BlankLine
{\bf Var:} epoch id $k = 0$\;
\While{Not all of samples in $D_u$ are labeled}{
    \BlankLine
    Run $E$ and $T_k$ on $D_u$ and get output probability vector set $P$ of $K$ classes\;
    \For{{\bf each} temporal adjacent dance phrases $d_{t-1}$ and $d_t$, probability $P(d_{t-1})$ and $P(d_{t-1})$ {\bf in} $D_u$, $P$}{
        \BlankLine
        Update $P(d_t)$: $P(d_t) \leftarrow P(d_t) M_k(d_{t-1} \rightarrow d_{t})$\;
        \BlankLine
    }
    Get pseudo labels $L$ based on re-scaled $P(d_t)$\;
    \BlankLine
    Initialize $D_u$'s subset $D_u'$ with a null set\;
    \For{{\bf each} dance movement $d$, label $l$, confidence $P(d)$ {\bf in} $D_u$, $L$, $P$}{
        \If{P(d) $>$ $\tau$}{
            Push $d$ and $l$ into $D_u'$\;
        }
    }
    Fine-tune the networks $T_k$ based on $D_l$ + $D_u'$ and get the new one $T_{k+1}$;
    \BlankLine
    Initialize $\mathbf{M}_{k+1}$ with a zero matrix\;
    \For{{\bf each} temporal adjacent phrases $d_{t-1}$ and $d_t$, and Top-1 confidences $P(d_{t-1})$ and $P(d_{t})$ {\bf in} $D_u$, $P$}{
        \BlankLine
        ${M}_{k+1}(d_{t-1} \rightarrow d_{t}) = M_k(d_{t-1} \rightarrow d_{t}) + P(d_{t-1}) P(d_{t}) $
        \BlankLine
    }
    Update the $\mathbf{M}_k$ with momentum: $\mathbf{M}_{k+1} \leftarrow \alpha \mathbf{M}_{k} + (1 - \alpha)\mathbf{M}_{k+1}$
    \BlankLine
    Update the epoch id: $k = k + 1$.
}
\KwResult{Output optimized $T^\star$ and $\mathbf{M}^\star$.}
\caption{Co-ascent learning algorithm}
\label{alg:co-ascent}
\end{algorithm}

\newpage
\subsection{Music and Dance Style Distribution on Datasets}
\begin{figure}[h]
  \includegraphics[width=0.6\linewidth]{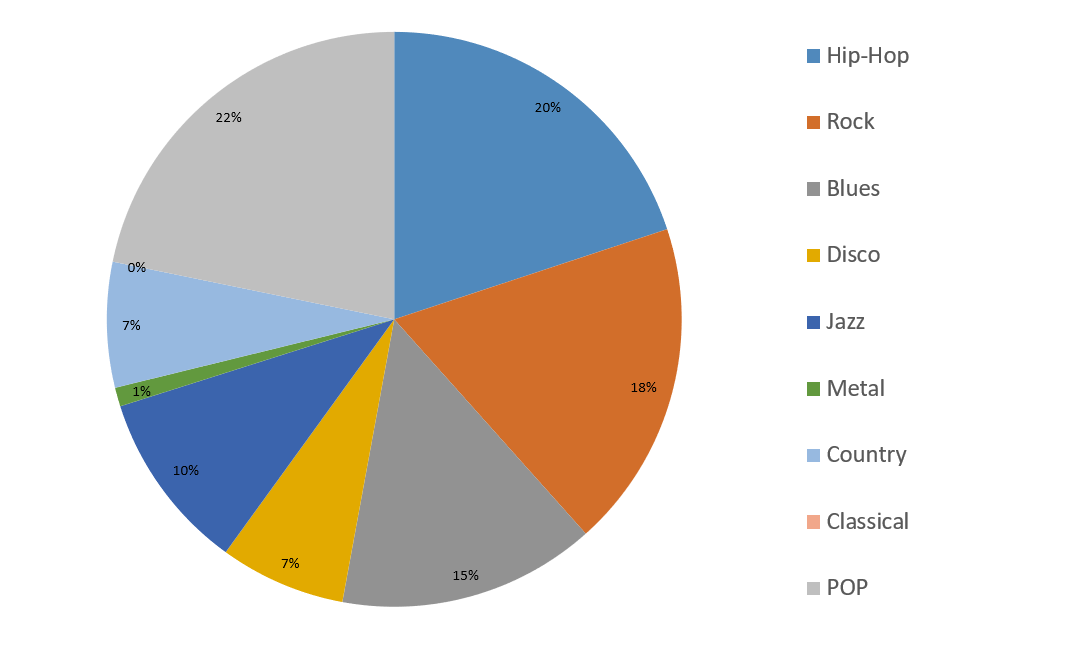}
  \centering
  \caption{Music style distribution on the small labeled dataset.}
  \label{supp_smallset}
\end{figure}

\begin{figure}[h]
  \includegraphics[width=0.6\linewidth]{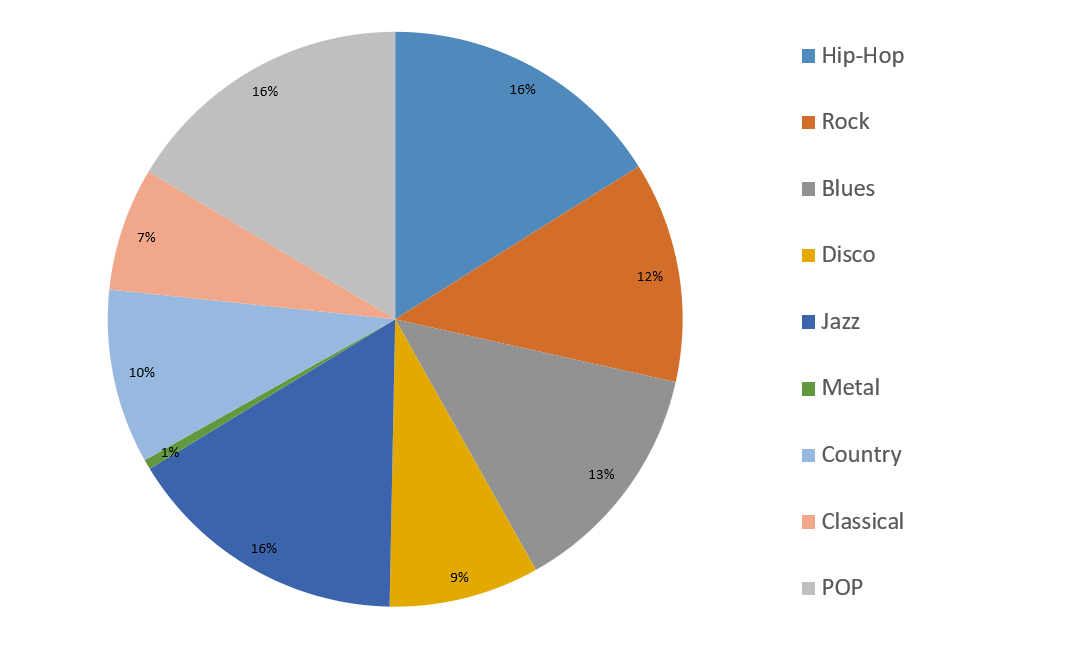}
  \centering
  \caption{Music style distribution on the large unlabeled dataset.}
  \label{supp_bigset}
\end{figure}

In Fig. \ref{supp_smallset} and Fig. \ref{supp_bigset}, we show the statistics of music style on two datasets, which can be roughly divided into 9 categories. For the dance phrases, we have dance styles including urban, jazz, hip-hop, popping, k-pop, locking, breaking and ACGN dance in our dataset to match the music styles of the small labeled dataset. It is worth mentioning that, based on the choreography theory, the matching of music and dance is flexible, thus our music style and dance style are not one-to-one correspondence.\par


\end{document}